\documentclass[letterpaper]{article} 
\usepackage[]{aaai23}  
\usepackage{times}  
\usepackage{helvet}  
\usepackage{courier}  
\usepackage[hyphens]{url}  
\usepackage{graphicx} 
\usepackage{natbib}  
\usepackage{caption} 
\usepackage{algorithm}
\usepackage{algorithmic}
\usepackage{booktabs}
\usepackage[utf8]{inputenc}
\usepackage{url}
\usepackage{booktabs}
\usepackage{amssymb}
\usepackage{bbding}
\usepackage{pifont}
\usepackage{wasysym}
\usepackage{utfsym}
\usepackage{fontawesome}
\usepackage{color}
\usepackage{newfloat}
\usepackage{listings}
\usepackage[colorlinks,linkcolor=red]{hyperref}
\DeclareCaptionStyle{ruled}{labelfont=normalfont,labelsep=colon,strut=off} 
\floatstyle{ruled}
\newfloat{listing}{tb}{lst}{}
\floatname{listing}{Listing}
\nocopyright 

\title{BEVStereo: Enhancing Depth Estimation in Multi-view 3D Object Detection with Dynamic Temporal Stereo}

\author{
    Yinhao Li\textsuperscript{\rm 1},
    Han Bao\textsuperscript{\rm 1},
    Zheng Ge\textsuperscript{\rm 2},
    Jinrong Yang\textsuperscript{\rm 3},
    Jianjian Sun\textsuperscript{\rm 2},
    Zeming Li\textsuperscript{\rm 2}
}
\affiliations{
    \textsuperscript{\rm 1}Institute of Computing Technology, Chinese Academy of Sciences,\\
    \textsuperscript{\rm 2}MEGVII Technology,
    \textsuperscript{\rm 3}Huazhong University of Science and Technology\\
    \{liyinhao, gezheng, yangjinrong, sunjianjian, lizeming\}@megvii.com, baohan20s@ict.ac.cn
}

\usepackage{bibentry}
\begin{document}
\pagestyle{plain}
\setcounter{page}{1}
\pagenumbering{arabic}
\maketitle

\begin{abstract}
Bounded by the inherent ambiguity of depth perception, contemporary camera-based 3D object detection methods fall into the performance bottleneck. Intuitively, leveraging temporal multi-view stereo (MVS) technology is the natural knowledge for tackling this ambiguity. However, traditional attempts of MVS are flawed in two aspects when applying to 3D object detection scenes: 1) The affinity measurement among all views suffers expensive computation cost; 2) It is difficult to deal with outdoor scenarios where objects are often mobile. To this end, we introduce an effective temporal stereo method to dynamically select the scale of matching candidates, enable to significantly reduce computation overhead. Going one step further, we design an iterative algorithm to update more valuable candidates, making it adaptive to moving candidates. We instantiate our proposed method to multi-view 3D detector, namely BEVStereo. BEVStereo achieves the new state-of-the-art performance (i.e., 52.5\% mAP and 61.0\% NDS) on the camera-only track of nuScenes dataset. Meanwhile, extensive experiments reflect our method can deal with complex outdoor scenarios better than contemporary MVS approaches. Codes have been released at \href{https://github.com/Megvii-BaseDetection/BEVStereo}{https://github.com/Megvii-BaseDetection/BEVStereo}.

\end{abstract}

\section{Introduction}
Due to the stability and inexpensive cost of vision sensors, camera-based 3D object detection has received extensive concern. Specially, the multi-view schemes~\cite{wang2022detr3d, huang2021bevdet, liu2022petr, li2022bevformer, huang2022bevdet4d, liu2022petrv2, li2022bevdepth} show significantly promising, and have made lots of breakthroughs. However, there is still a substantial performance gap compared with LiDAR-based approaches~\cite{lang2019pointpillars, yan2018second, yin2021center}, since it exposes a notoriously ill-posed issue for perceiving depth.


Contemporary multi-view detectors~\cite{huang2021bevdet, huang2022bevdet4d, li2022bevdepth} predict a discrete depth distribution for each point of the field of view (FOV), which enables to project features from image representation to BEV map. The unified BEV map is the key to learning harmonious results since the overlap regions of adjacent views represent more complete to directly forecast results. Such sweetness is hard to be enjoyed by the monocular-based detector~\cite{wang2021fcos3d}, as a post-processing strategy is needed to remove repetitive and low-quality 3D boxes in overlap areas. 


The above paradigm is based on an important preconceived assumption, i.e., the perceived depth distribution in FOV needs accurate enough. However, most of them perceive depth by only feeding into single-frame images, which is actually an ill-posed solution~\cite{huang2021bevdet, huang2022bevdet4d, li2022bevdepth}. Several studies~\cite{yao2018mvsnet, xue2019mvscrf, bae2022multi} point out that predicting depth needs \textbf{multi-view stereo} condition, which requires images from different views to construct cost volume. Fortunately, the automatic driving scenario is often processed in a continuous time sequence, enabling us to leverage temporal views for constructing multi-view stereo. 

To carry out the traditional temporal stereo technology like \cite{yao2018mvsnet} is non-trivial in automatic driving scenarios, which manifests in two aspects:



\begin{enumerate}
\item \textbf{Large memory cost.} When we replace the depth module in BEVDepth with a basic temporal stereo method~\cite{yao2018mvsnet}, the memory cost grows to 3.5 times that of BEVDepth despite bringing a 1.6 percent promotion on NDS, making it a tremendous burden to apply it to a detection task;

\item \textbf{Failing to reason the depth of moving objects and static ego vehicle cases.} Temporal stereo approaches are unable to handle several situations~\cite{wang2022monocular} like a static ego vehicle and moving objects since the parallax angle tends to 0 if ego vehicle is static and the stereo is unable to match if the object is moving. However, after statistics in nuScenes scene, over 10\% of the frames' ego vehicles are static, while approximately 25\% of the objects are moving. Therefore, these two shortcomings limit its application to autonomous driving scenarios.
\end{enumerate}

MVS methods~\cite{wang2022monocular, wang2022mv} expose that the majority of the computational memory cost is associated with constructing cost volume due to its calculation procedure of dense similarity. It naturally motivates us to construct a \emph{sparse cost volume} for cutting computational memory. To this end, we propose a dynamic mechanism to sample a small number of reference candidate features for building cost volume instead of all ones along the depth axis. It is implemented by predicting two modeling parameters, i.e., depth center $\mu$ and depth range $\sigma$. This far, it can significantly reduce computational memory. Going into one step, we introduce a parameter evolution method for $\mu$ and $\sigma$, which is carried out by applying the EM algorithm to update the modeling parameters $\mu$ and $\sigma$. With the evolution technique, it is possible to continuously improve reference candidate features that are more important for cost volume while adjusting to situations including moving objects and stationary ego vehicles. This insight is similar to MaGNet~\cite{bae2022multi}, but it not only fails to deal with complex outdoor situations but introduces redundantly learnable parameters to update $\mu$ and $\sigma$. Finally, we also introduce an advanced variant of Circle NMS~\cite{yin2021center}, which takes objects' size into account for better removing duplicate 3D boxes.

We instantiate our proposed methods to advanced BEVDepth~\cite{li2022bevdepth}, namely \textbf{BEVStereo}. By conducting comprehensive experiments on nuScence benchmark~\cite{caesar2020nuscenes}, it shows significant improvements in the 3D object detection task. In conclusion, the contributions of this work are as three-fold as follows:

\begin{itemize}
\item We point out that the MVS technology is a promising method for tackling the ill-posed issue of depth perception in camera-based 3D object detection task. But it exposes two fatal flaws in the automatic driving scenarios, i.e., either large memory cost issue or moving objects and static ego vehicles.

\item We introduce a dynamic temporal stereo technique, which can save extreme memory cost to construct cost volume. Moreover, a parameter evolution algorithm is proposed to tackle moving and noisy features of objects. 


\item BEVStereo improves mAP and NDS by 1.7\% and 1.7\% on nuScenes dataset, while achieving the new SOTA performance on the camera-only track. Extensive experiments verify that our approach can effectively be adapted to moving objects and static ego vehicles.


\end{itemize}


\section{Related Work}
\subsection{Single-view 3D Object Detection}
Many approaches have made their effort on predicting objects directly from single images. For the purpose of 3D object detection, Cai et al.~\cite{cai2020monocular} calculates the depth of the objects by integrating the height of the objects in the image with the height of the objects in the real world. Based on FCOS~\cite{tian2019fcos}, FCOS3D~\cite{wang2021fcos3d} extends it to 3D object detection by changing the  classification branch and regression branch which predicts 2D and 3D attributes at the same time. M3D-RPN~\cite{brazil2019m3d} treats mono-view 3D object detection task as a stand-alone 3D region proposal network, narrowing the gap between LiDAR-based approaches and camera-based methods. D$^4$LCN~\cite{ding2020learning} replaces 2D depth map with pseudo LiDAR representation to better present 3D structure. DFM~\cite{wang2022monocular} integrates temporal stereo to mono-view 3D object recognition, improving the quality of depth estimation while minimizing the negative effects of difficult situations that temporal stereo is unable to handle.
\subsection{Multi-view 3D Object Detection}
Current multi-view 3D object detectors can be divided into two schemas: LSS-based ~\cite{philion2020lift} schema and transformer-based schema. 

BEVDet ~\cite{huang2021bevdet} is the first study that combines LSS and LiDAR detection head which uses LSS to extract BEV feature and uses LiDAR detection head to propose 3D bounding boxes. By introducing previous frames, BEVDet4D~\cite{huang2022bevdet4d} acquires the ability of velocity prediction. To reduce memory usage, M$^2$BEV~\cite{xie2022m}  decreases the learnable parameters and achieves high efficiency on both inference speed and memory usage. BEVDepth~\cite{li2022bevdepth} uses LiDAR to generate depth GT for supervision and encodes camera intrinsic and extrinsic parameters to enhance the model's ability of depth perception.

DETR3D~\cite{wang2022detr3d} extends DETR ~\cite{carion2020end} into 3D space, using transformer to generate 3D bounding boxes. Based on DETR, PETR~\cite{liu2022petr} and PETRV2~\cite{liu2022petrv2} adds position embedding onto it. BEVFormer~\cite{li2022bevformer} uses deformable transformer to extract features from images and uses cross attention to link the feature between frames for velocity prediction. 

\subsection{Depth Estimation}
Based on the number of images used for depth estimation, depth estimation methods can be divided into single-view depth estimation and multi-view depth estimation.

Although predicting depth from a single image is obviously ill-posed, it is still possible to estimate some of the depth of the objects by using the context as a signal. Therefore, many approaches~\cite{bhat2021adabins, eigen2015predicting, eigen2014depth, fu2018deep} use CNN method to predict depth.

For the task of multi-view depth estimation, Constructing cost volume is an effective way to predict depth~\cite{mvs_survery, aa_rmvsnet, bidirectional}. MVSNet~\cite{yao2018mvsnet} is the first research that uses cost volume for depth estimation. RMVSNet~\cite{yao2019recurrent} reduces memory cost by introducing GRU module. MVSCRF~\cite{xue2019mvscrf} adds CRF module onto MVSNet. PointMVSNet~\cite{chen2019point} uses point algorithm to optimize the regression of depth estimation. Cascade MVSNet~\cite{gu2020cascade} uses cascade structure, making it able to use large depth range and a small amount of depth intervals. Fast-MVSNet~\cite{yu2020fast} uses sparse cost volume and Gauss-Newton layer to speed up MVSNet. Wang et al.~\cite{wang2021patchmatchnet} use adaptive patchmatch and multi-scale fusion to achieve good performance while mataining high efficiency. Bae et al.~\cite{bae2022multi} introduce MaGNet to better fuse single-view depth estimation and multi-view depth estimation.

\section{Method}
BEVStereo is a stereo-based multi-view 3D object detector.
By applying our temporal stereo technique, it is able to handle complex outdoor scenarios while maintaining memory efficiency. We also propose a size-aware circle NMS approach to improve the proposal suppression process.

\subsection{Preliminary Knowledge}
\paragraph{Multi-view 3D object detection}
LSS-based~\cite{philion2020lift} multi-view 3D object detectors currently include four components: an image encoder to extract the image features, a depth module to generate depth and context, then outer product them to get point features, a view transformer to convert the feature from camera view to the BEV view, and a 3D detection head to propose the final 3D bounding boxes.
\paragraph{Temporal stereo methods to predict depth}
MVS-based~\cite{yao2018mvsnet} methods predict depth by constructing cost volume. For every pixel on the reference feature, they initially put forth a number of candidates along the depth axis. They next convert these candidates from reference to source using a homography warping operation in order to retrieve the relevant source feature and create the cost volume. After cost volume is constructed. For the purpose of predicting the confidence of each depth candidate, 3D convolution is performed to regularize the cost volume.

\subsection{Dynamic Temporal Stereo}
\begin{figure*}[t]
\includegraphics[width=0.9\textwidth]{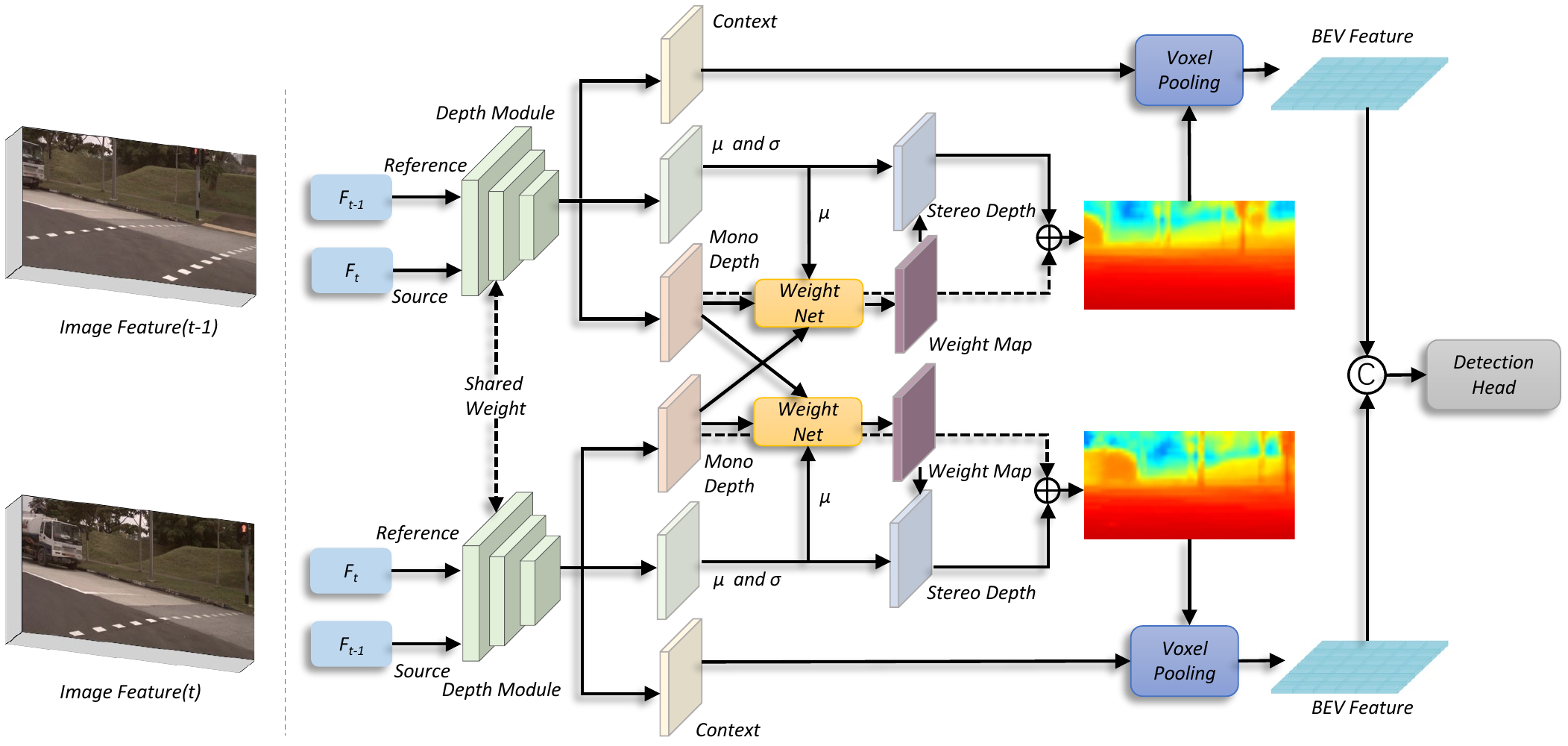}
\centering
\caption{Framework of BEVStereo. The Depth Module uses the image feature of the reference frame and source frame as input to generate $\mu$, $\sigma$, context, and mono depth. Stereo depth is produced using $\mu$ and $\sigma$. Weight Net uses $\mu$ and the mono depth of two frames to create a weight map that is applied to the stereo depth. Mono depth and weighted stereo depth are accumulated together to create the final depth. BEV Feature is produced when context is combined with it and is used by the detecting head.}
\label{fig:bevdepth}
\end{figure*}

Based on BEVDepth ~\cite{li2022bevdepth}, BEVStereo changes the way of generating depth prediction. Instead of predicting depth from a single image, BEVStereo predicts both depth from single feature (mono depth) and depth from temporal stereo (stereo depth). For mono depth, we directly predict depth prediction, which is the same as BEVDepth. For stereo depth, we firstly predict depth center ($\mu$) and depth range ($\sigma$), then $\mu$ and $\sigma$ are used to generate depth distribution. Additionally, Weight Net is used to create a weight map that will be applied on stereo depth. Mono depth and weighted stereo depth are combined to get the final depth. Our framework overview is illustrated in Fig.~\ref{fig:bevdepth}. 

\paragraph{Depth Module}
\begin{figure}[t]
\includegraphics[width=0.4\textwidth]{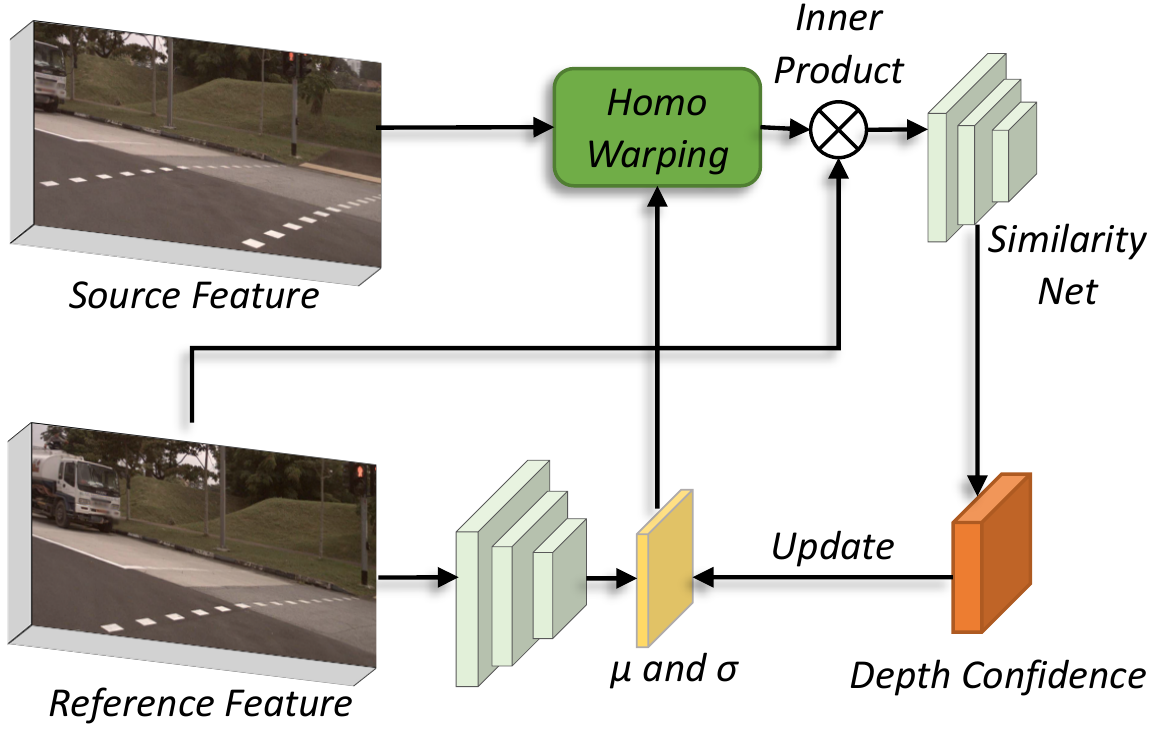}
\centering
\caption{Iterative process of $\mu$ and $\sigma$. The initial $\mu$ and $\sigma$ are generated using feature of the reference frame as input. For each round of iteration, $\mu$ and $\sigma$ are used for homography warping to fetch the source feature. Similarity Net takes the inner product results of warpped source feature and reference feature as input to generate depth confidence which is used to update $\mu$ and $\sigma$.}
\label{fig:depth_module}
\end{figure}
Our Depth Module simultaneously predicts  mono depth, $\mu$, $\sigma$ and context. After iterating $\mu$ and $\sigma$ by our EM method, they are used to generate the stereo depth. The process of iterating $\mu$ and $\sigma$ is illustrated in Fig.~\ref{fig:depth_module}.

We choose to estimate $\mu$ and $\sigma$, which stand for the depth center and depth range of the cost volume.
Compared to other stereo-based methods of splitting bins along the depth dimension~\cite{yao2018mvsnet, wang2022mv}, our method can dynamically choose the search area while also lowering the number of candidates. After estimating $\mu$ and $\sigma$ of the reference frame, we can dynamically select candidates for each pixel based on the depth center and range of cost volume and obtain the depth of these candidates. These candidates are used for homography warping operation to fetch the feature from source frame, as illustrated in Equ.~\ref{equ: homowarping}, where $P$ denotes the coordinate of the point, $D$ denotes the depth of the candidate, $src$ denotes source frame, $ref$ denotes reference frame, $M_{ref2src}$ denotes the transformation matrix from the reference frame to source frame and $K$ denotes the intrinsic matrix. The reference feature and the warpped source feature are used to construct cost volume. Similarity Net is followed to predict the confidence score of all candidates.

\begin{equation}
\label{equ: homowarping}
    P_{src}[u\cdot z, v \cdot z, z] = K \times M_{ref2src} \times K^{-1} \times (D \cdot P_{ref}[u, v, 1])
\end{equation}

Inspired by the EM algorithm, We attempt to make the expectation of $\mu$ closer to the depth gt during the iteration process. Since we compute each point's confidence after sampling a number of points close to $\mu$, it is only natural that we use this knowledge to further our objectives. As a result, we update $\mu$ using the weight sum method, which causes $\mu$ to become the expectation of the sample points for each iteration. The update rule is illustrated in Eq.~\ref{equ: weight-sum}, where $D_i$ denotes the depth of the $i$th candidate and $P_i$ denotes the probability of the $i$th candidate. When facing cases like static ego vehicle and moving objects, all candidates share the same low probability since it is hard to find the best match point on the source feature, $\mu$ is able to maintain its value by using the weight sum technique. For other scenarios, the value of $\mu$ will approach the true depth value in the process of iteration. Surprisingly, we discover that when $\mu$ and mono depth are trained together, the quality of initial $\mu$ is also enhanced under the direction of mono depth. Therefore, in all kinds of scenarios, our dynamic temporal stereo approach can improve depth prediction. As $\mu$ is being updated in the process of iteration, it is also critical to find the suitable $\sigma$ to set the searching range. In accordance with existing information, the searching range should be reduced when the confidence of $\mu$ is high and expanded when it is low, we update $\sigma$ following Equ.~\ref{equ: update-sigma} where $P_{\mu}$ denotes the confidence of $\mu$. Without introducing any learnable parameters, the search range is optimized during iteration. 

To prevent the scenario where the projected $\mu$ is far from the depth gt, making it difficult to optimize $\mu$ during iteration. we divide the depth into different ranges and use our iteration technique in each split range. After the iteration process is finished, the depth map is generated following  Equ.~\ref{equ: depth-map} where P denotes the computed depth confidence and D denotes the depth of the split bins along the depth axis for each pixel. 

\begin{equation}
\label{equ: weight-sum}
    \mu = \sum_{i=1}^n D_i \cdot P_i,
\end{equation}

\begin{equation}
\label{equ: update-sigma}
    \sigma_{new} =  \frac{\sigma_{old}}{2 \cdot P_{\mu}},
\end{equation}

\begin{equation}
\label{equ: depth-map}
    P = exp(-\frac{1}{2} \cdot (\frac{D - \mu}{\sqrt{\sigma}})^2).
\end{equation}
\paragraph{Weight Net}
Even while the temporal stereo is capable of accurately predicting depth, there are still some areas where it is unreliable because some reference feature points do not correlate to positions on source feature. Therefore, we introduce Weight Net to better combine mono depth and stereo depth. To do this, we apply the same homography warping operation to fetch the mono depth of the source frame, using $\mu$ as the depth. A similarity net is then applied to the warped mono depth from the source frame and the mono depth from the reference frame to construct the weight map.

\subsection{Size-aware Circle NMS}
\begin{figure}[t]
\includegraphics[width=0.3\textwidth]{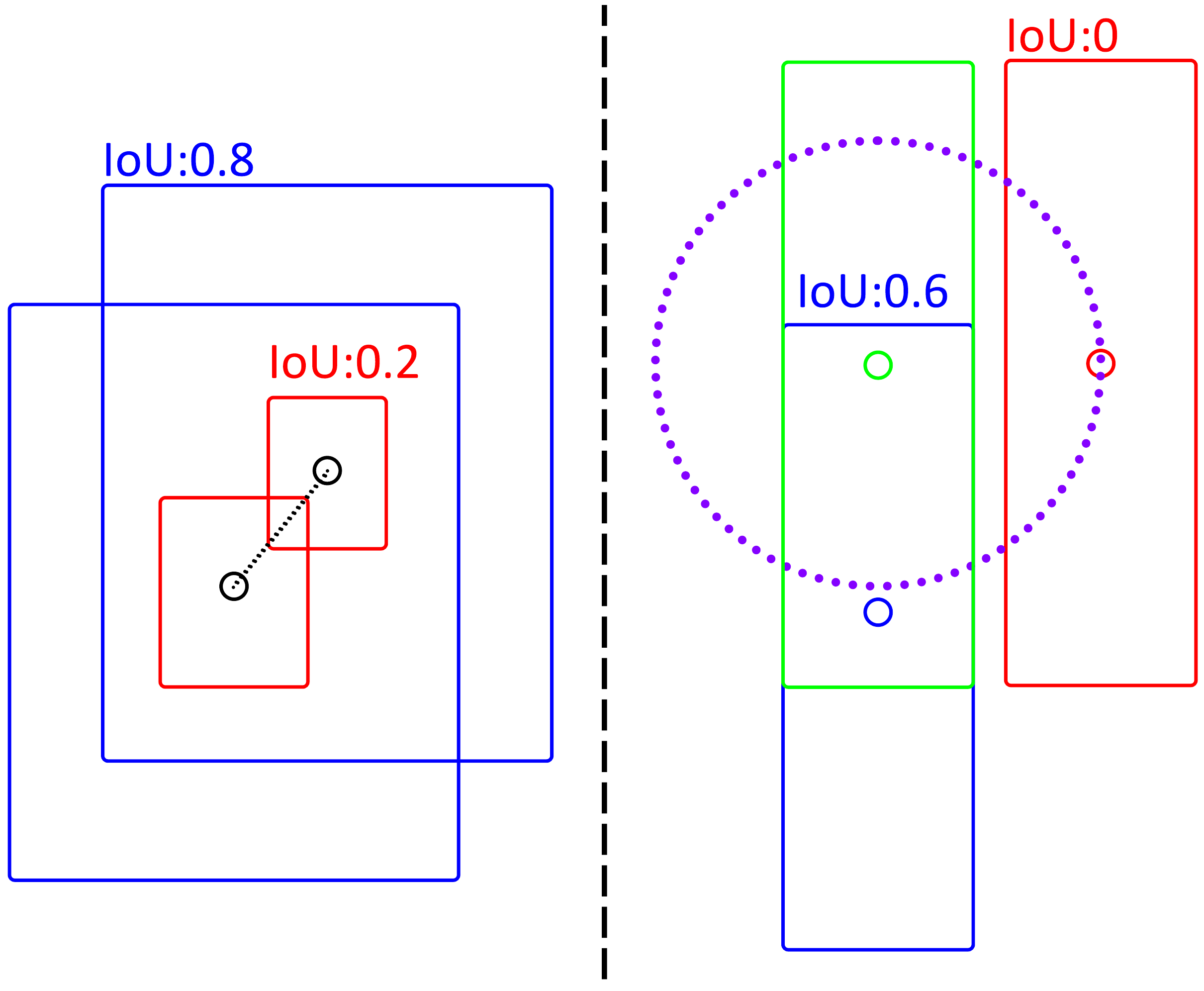}
\centering
\caption{Drawbacks of circle NMS. In the left part of the figure, despite having distinct IoUs, the blue boxes and red boxes share the same center distance as long as their centers coincide. In the right part of the figure, when the green box has the highest score, the red box is more likely to be suppressed since its center is closer to the green box's center which goes against our common sense.}
\label{fig:circle-nms}
\end{figure}

The distance between the centers of two bounding boxes is used by circle NMS ~\cite{yin2021center} function as a criterion for suppression. Circle NMS achieves excellent efficiency and good performance by bypassing the difficult process of computing rotated IoU of bouding boxes. However, ignoring the size of boxes will result in two drawbacks as illustrated in Fig.~\ref{fig:circle-nms}: 1) No matter how closely the boxes overlap, the NMS algorithm yields the same output as long as the box centers are fixed. 2) When boxes are placed differently, boxes with 0 IoU may be removed while boxes with high IoU are kept. 

We propose size-aware circle NMS, which avoids computing rotated IoU while taking into consideration the size of the boxes. We separate the distance of two bounding boxes' centers into x axis and y axis. We use $x_{thre}$ and $y_{thre}$ as threholds of x axis and y axis, which are computed following Equ.~\ref{equ: xthre} and Equ.~\ref{equ: ythre}, where $\theta$ denotes the orientation, $w$ denotes the hyper parameter of scale factor, $d_x$ denotes the length of the box and $d_y$ denotes the width of the box. The box will be suppressed when the distance in x axis is smaller than $x_{thre}$ and distance in y axis is smaller than $y_{thre}$. By applying size-aware circle NMS, the blue box with a lower score will be suppressed in scenarios like the left portion of Fig~\ref{fig:circle-nms} because it has a greater $x_{thre}$ and $y_{thre}$. The blue box will be suppressed in scenarios like the right portion of Fig.~\ref{fig:circle-nms} because the distances in the x and y axes are more likely to be smaller than $x_{thre}$ and $y_{thre}$ in the mean time.
\begin{equation}
\label{equ: xthre}
    x_{thre} = w \cdot (sin\theta_1 \cdot d_{x1} + cos\theta_1 \cdot d_{y1} + sin\theta_2 \cdot d_{x2} + cos\theta_2 \cdot d_{y2}).
\end{equation}

\begin{equation}
\label{equ: ythre}
    y_{thre} = w \cdot (sin\theta_1 \times d_{y1} + cos\theta_1 \cdot d_{x1} + sin\theta_2 \cdot d_{y2} + cos\theta_2 \cdot d_{x2}).
\end{equation}

\begin{figure}[b]
\includegraphics[width=0.45\textwidth]{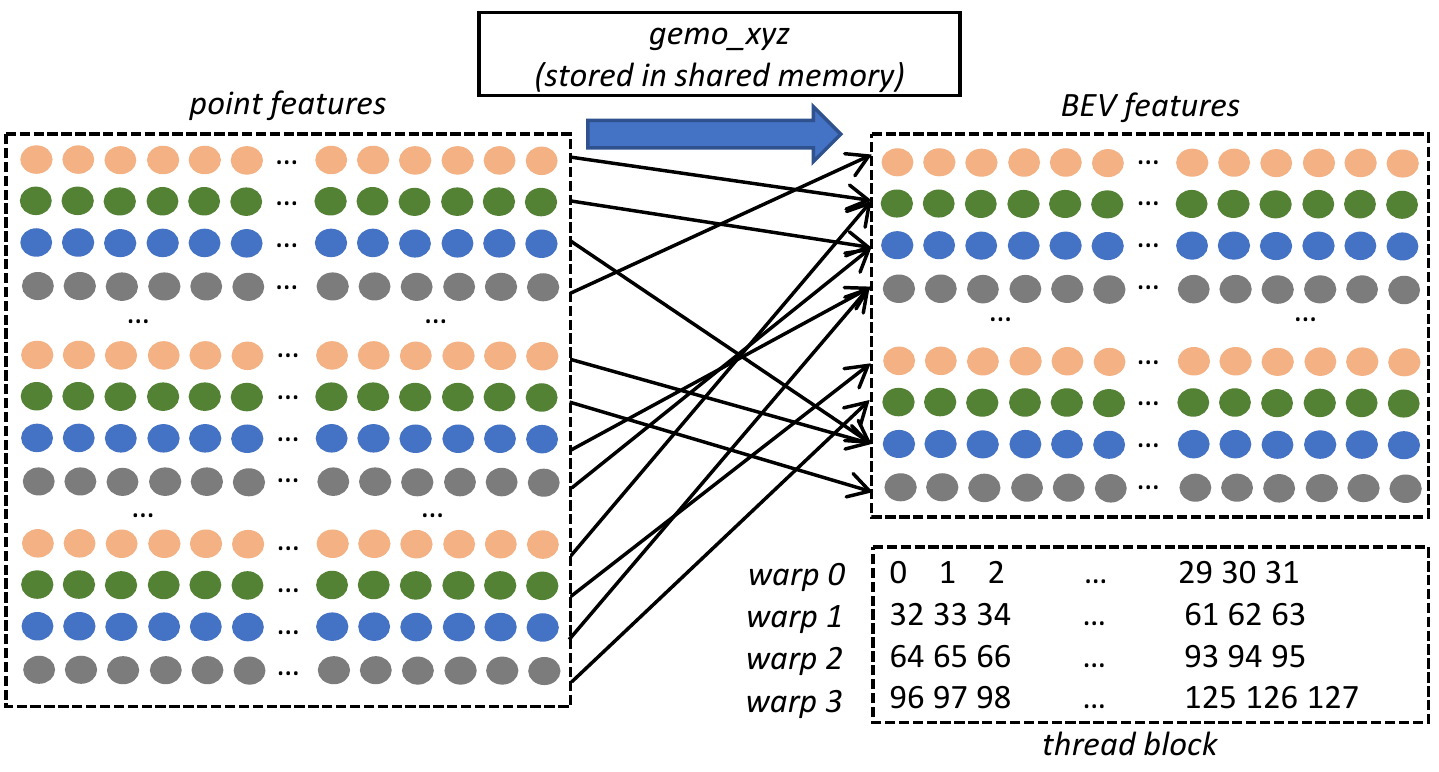}
\centering
\caption{Thread mapping of point features to BEV features. Based on the point coordinates, the point features are atomically accumulated into the corresponding BEV features. Each thread block loads the point coordinates it is responsible for into the shared memory.}
\label{fig:mapping}
\end{figure}

\begin{table*}
\centering
\scalebox{0.8}{
\begin{tabular}{c|c|c|ccccccc}
\toprule
 \textbf{Method} & \textbf{WN}  & \textbf{mAP}$\uparrow$  & \textbf{mATE}$\downarrow$ & \textbf{mASE}$\downarrow$  & \textbf{mAOE}$\downarrow$ & \textbf{mAVE}$\downarrow$ & \textbf{mAAE}$\downarrow$ & \textbf{NDS}$\uparrow$ \\
\midrule
    BEVDepth & & 32.7 &  70.1 & 27.7 & 55.6 & 55.8 & 21.4 & 43.3  \\ \midrule
 BEVStereo &  & 34.5 & 66.5 & 27.9 & 52.9 & 55.0 & 23.6 & 44.7   \\ 
 BEVStereo & \checkmark & 34.6 & 65.3 & 27.4 & 53.1 & 51.6 & 23.0 & 45.3   \\
\bottomrule
\end{tabular}}
\caption{Detection results on the nuScenes \emph{val} set. WN denotes Weight Net.}
\label{tab:performance-analysis-1}
\end{table*}

\begin{table}[t]
\centering
\scalebox{0.8}{
\begin{tabular}{c|ccccc} 
\toprule
\textbf{Method} &\textbf{SILog}$\downarrow$   & \textbf{Abs Rel}$\downarrow$ & \textbf{Sq Rel}$\downarrow$     & \textbf{log10}$\downarrow$  & \textbf{RMSE}$\downarrow$  \\
\midrule
   BEVDepth & 21.74 & 0.155 & 1.223 & 0.060 & 5.269     \\
   \midrule
 BEVStereo & 21.74 & 0.152 & 1.206 & 0.059 & 5.246    \\
\bottomrule
\end{tabular}}
\caption{Evaluation of depth prediction on the nuScenes \emph{val} set.}
\label{tab:ablation-depth}
\end{table}

\begin{table}[t]
\centering
\scalebox{0.8}{
\begin{tabular}{c|c|ccccc} 
\toprule
\textbf{Method} & \textbf{Iter} &\textbf{TH=0.5}   & \textbf{TH=1}& \textbf{TH=2}     & \textbf{TH=4}  \\
\midrule
   BEVDepth & & 28.32 & 46.10 & 60.37 & 	71.18    \\
   BEVDepth + MVS & & 27.67 & 46.40 & 59.99 & 71.26     \\
   \midrule
 BEVStereo & & 29.79 & 49.26 & 61.53 & 72.10    \\
 BEVStereo & \checkmark& 29.40 & 48.97 & 61.53 & 72.27    \\
\bottomrule
\end{tabular}}
\caption{Recall results on the nuScenes \emph{val} set. Only boxes with velocity higher than 1m/s are maintained for analysis.  BEVDepth + MVS denotes replacing depth module in BEVDepth with MVS approach. Different thresholds are utilized depending on the distance between boxes' center. Iter denotes whether to iterate $\mu$ during the inference stage.}
\label{tab:analysis-velocity}
\end{table}

\begin{table}[h!]
\centering
\scalebox{0.8}{
\begin{tabular}{c|ccccc} 
\toprule
\textbf{Method} &\textbf{TH=0.5}   & \textbf{TH=1}& \textbf{TH=2}     & \textbf{TH=4}  \\
\midrule
   BEVDepth & 32.80 & 53.58 & 70.00 & 80.89    \\
   BEVDepth + MVS & 33.61 & 54.23 & 69.89 & 80.57     \\
   \midrule
 BEVStereo & 33.90 & 54.79 & 70.51 & 81.01    \\
\bottomrule
\end{tabular}}
\caption{Recall results on the nuScenes \emph{val} set. Only boxes with velocity lower than 1m/s are maintained for analysis.}
\label{tab:analysis-static}
\end{table}

\begin{table}
\centering
\scalebox{0.8}{
\begin{tabular}{c|c|ccccccc}
\toprule
 \textbf{Method} & \textbf{Iter} & \textbf{mAP}$\uparrow$  & \textbf{mATE}$\downarrow$ & \textbf{NDS}$\uparrow$ \\
\midrule
    BEVDepth & & 32.73 & 73.47 & 44.14  \\
  BEVDepth + MVS & & 31.55 & 78.06 & 43.21   \\
  \midrule
 BEVStereo &  & 33.12 & 63.01 & 46.68
   \\ 
 BEVStereo & \checkmark & 33.76 & 63.49 & 46.76
   \\ 
\bottomrule
\end{tabular}}
\caption{Detection results on the nuScenes \emph{val} set. Only frames with ego vehicles moving at speeds less than 1 m/s are employed for evaluation.}
\label{tab:analysis-ego-vehicle}
\end{table}

\begin{table}[h!]
\centering
\scalebox{0.8}{
\begin{tabular}{c|c|ccc}
\toprule
 \textbf{Method} & \textbf{Memory} & \textbf{mAP}$\uparrow$  & \textbf{mATE}$\downarrow$ & \textbf{NDS}$\uparrow$ \\
\midrule
    BEVDepth &  6.49GB & 32.7 &  70.1 & 43.3  \\
  BEVDepth + MVS & 24.04GB & 34.7 & 67.1 & 44.9   \\
  \midrule
 BEVStereo & 8.01GB & 34.6 & 65.3 & 45.3   \\ 
\bottomrule
\end{tabular}}
\caption{Memory usage and detection results of BEVDepth, BEVDepth with MVS and BEVStereo.}
\label{tab:memory-efficiency}
\end{table}

\begin{table}[h!]
\centering
\scalebox{0.8}{
\begin{tabular}{c|ccc}
\toprule
 num\_iter & \textbf{mAP}$\uparrow$  & \textbf{mATE}$\downarrow$ & \textbf{NDS}$\uparrow$ \\
\midrule
0 & 32.7 & 67.4 & 43.9  \\
1 & 33.1 & 67.0 & 44.2   \\ 
2 & 34.1 & 65.9 & 45.0  \\ 
3 & 34.6 & 65.3 & 45.3  \\ 
\bottomrule
\end{tabular}}
\caption{Detection results on the nuScenes \emph{val} set. num\_iter denotes the number of iterations for $\mu$.}
\label{tab:iteration}
\end{table}

\begin{table}[h!]
\centering
\scalebox{0.8}{
\begin{tabular}{c|c|ccc}
\toprule
 \textbf{Method} & CA & \textbf{mAP}$\uparrow$  & \textbf{mATE}$\downarrow$ & \textbf{NDS}$\uparrow$ \\
\midrule
    circlenms & &  34.6 & 65.3 & 45.3  \\
    circle-nms & \checkmark&  24.9 & 80.6 & 38.0  \\
    \midrule
  size-aware-circlenms & & 35.1 & 64.7 & 45.6   \\ 
  size-aware-circlenms & \checkmark&   33.3 & 64.1 & 45.0 \\ 
\bottomrule
\end{tabular}}
\caption{Detection results on the nuScenes \emph{val} set. CA denotes class-agnostic. All results are conducted under the best hyper parameters.}
\label{tab:circlenms}
\end{table}

\section{Experiment}
In this section, we first describe the experimental settings that we employ before going into the specifics of our implementation strategy. Experiments involving heavy ablation are carried out to confirm the efficacy and validity of BEVStereo.

\subsection{Experimental Settings}
\paragraph{Dataset and evaluation metrics}
We decide to run our experiments on the nuScenes~\cite{caesar2020nuscenes} dataset. For training, we use LiDAR and image data, but we only use image data for inference. In the case of image data, the key frame image and the furthest sweep connected to it are used, whereas in the case of LiDAR data, only the key frame data is used. We assess the results of our method using detection and depth metrics. Memory usage is also used to assess the effectiveness of our method. To be more specific, we report the mean Average Precision (mAP), nuScenes Detection Score (NDS), mean Average Translation Error (mATE), mean Average Scale Error (mASE), mean Average Orientation Error (mAOE), mean Average Velocity Error (mAVE), and mean Average Attribute Error (mAAE). We follow the established evaluation procedures for the depth estimation task~\cite{deptheval}, reporting scale invariant logarithmic error (SILog), mean absolute relative error (Abs Rel), mean squared relative error (Sq Rel), mean log10 error (log10), and root mean squared error (RMSE) to assess our approach.

\paragraph{Implementation details}
We implement BEVStereo based on BEVDepth~\cite{li2022bevdepth}. The feature map we employ for building the cost volume has a downsampling rate of 4 while the depth feature's final form remains unchanged. The MVS~\cite{yao2018mvsnet} approach is applied to replace the depth module in BEVDepth with the same input resolution and output resolution in order to fairly demonstrate the effectiveness of our method. The learning rate is set to 2e-4, the EMA technique is also used, and AdamW~\cite{adamw} is used as the optimizer. During training, we use both image and BEV data augmentation.
\subsection{Analysis}
We perform numerous experiments to examine the mechanism of BEVStereo in order to better understand how it works. We choose BEVDepth ~\cite{li2022bevdepth} as baseline, we also implement MVSNet~\cite{yao2018mvsnet} on BEVDepth as a comparison to  show the distinct benefit that BEVStereo provides,  detection results and recall results are used for comparison.
\paragraph{Memory analysis}
We keep track of memory usage and detection results to demonstrate how effectively we use our memory. We also monitor the same matrics for the MVS-based~\cite{yao2018mvsnet} approach for fair comparison.

As illustrated in Tab.~\ref{tab:memory-efficiency}, BEVStereo increases the metrics on mAP, mATE, and NDS considerably at the expense of adding little memory consumption. When compared to using MVS~\cite{yao2018mvsnet} on BEVDepth~\cite{li2022bevdepth}, BEVStereo considerably reduces memory usage while boosting performance.

\paragraph{Performance analysis}
To begin with, we demonstrate the performance comparison under the nuScenes~\cite{caesar2020nuscenes} evaluation metrics. As shown in Tab.~\ref{tab:performance-analysis-1}, Our BEVStereo outperforms BEVDepth on mAP, mATE and NDS. Tab.~\ref{tab:ablation-depth} shows that the accuracy of depth estimation is improved by introducing our design.

We assess the performance of BEVStereo under challenging conditions such as moving objects, and static ego vehicles in order to show how well it adapts to complicated outdoor environments. Tab.~\ref{tab:analysis-velocity} demonstrates that BEVStereo still has the ability to improve performance even while MVS approach fails when dealing with moving objects. The static objects, which make up the majority of MVS schema's contribution, are also used to evaluate our method. As shown in Tab.~\ref{tab:analysis-static}, BEVStereo's ability of perceiving static objects is even higher than BEVDepth with MVS. We choose frames whose ego vehicle has a low velocity for evaluation since MVS cannot handle situations when this occurs. As can be seen in Tab.~\ref{tab:analysis-ego-vehicle}, BEVStereo still improves performance even when MVS fails in these conditions. It is important to note that BEVStereo still produces the similar results when faced with circumstances like moving objects and static ego vehicles if $\mu$ is not updated during the inference step. This demonstrates that our schema is capable of guiding the Depth Module to produce better $\mu$ and maintaining the initial prediction of $\mu$ in the face of these eventualities.

\begin{table*}[!t]
\centering
\resizebox{0.98\textwidth}{!}{
\begin{tabular}{l|c|cccccc|c}
\toprule
\textbf{Method}                                   & \textbf{Modality} & \textbf{mAP}$\uparrow$  & \textbf{mATE}$\downarrow$ & \textbf{mASE}$\downarrow$  & \textbf{mAOE}$\downarrow$ & \textbf{mAVE}$\downarrow$ & \textbf{mAAE}$\downarrow$ & \textbf{NDS}$\uparrow$ \\
\midrule
CenterPoint                         & L        & 0.564 & - & - & -   & - & - & 0.648  \\ \midrule
FCOS3D~\cite{wang2021fcos3d}                                   & C        & 0.358 & 0.690 & 0.249 & 0.452   & 1.434 & 0.124 & 0.428  \\
DETR3D~\cite{wang2022detr3d}                                   & C        & 0.412 & 0.641 & 0.255 & 0.394   & 0.845 & 0.133 & 0.479  \\
BEVDet-Pure~\cite{huang2021bevdet}                              & C        & 0.398 & 0.556 & 0.239 & 0.414   & 1.010 & 0.153 & 0.463  \\
BEVDet-Beta                              & C        & 0.422 & 0.529 & 0.236 & 0.396   & 0.979 & 0.152 & 0.482  \\
PETR~\cite{liu2022petr} & C        & 0.434 & 0.641 & 0.248 & 0.437   & 0.894 & 0.143 & 0.481  \\
PETR-e                                   & C        & 0.441 & 0.593 & 0.249 & 0.384   & 0.808 & 0.132 & 0.504  \\
BEVDet4D~\cite{huang2022bevdet4d}                                 & C        & 0.451 & 0.511 & 0.241 & 0.386   & 0.301 & 0.121 & 0.569  \\
BEVFormer~\cite{li2022bevformer}                                & C        & 0.481 & 0.582 & 0.256 & 0.375   & 0.378 & 0.126 & 0.569  \\ 
PETRv2~\cite{liu2022petrv2}                                & C        & 0.490 & 0.561 & 0.243 & 0.361   & 0.343 & 0.120 & 0.582  \\ 

BEVDepth~\cite{li2022bevdepth}                               & C        & 0.503 & 0.445 & 0.245 & 0.378 & 0.320 & 0.126 & 0.600 \\
\midrule
BEVStereo                                 & C        & 0.525 & 0.431 & 0.246 & 0.358 & 0.357 & 0.138 & 0.610 \\

\bottomrule
\end{tabular}}
\caption{Comparison on the nuScenes \emph{test} set. L denotes LiDAR and C denotes camera.}
\label{tab:test}
\end{table*}

\begin{table}[t]
\centering
\vspace{-2mm}
\scalebox{0.6}{
\setlength{\tabcolsep}{1.5pt}
\begin{tabular}{l|c|c|cc} 
\toprule
\textbf{Method}             & \textbf{Resolution} & \textbf{Modality} & \textbf{mAP}$\uparrow$  & \textbf{NDS}$\uparrow$ \\
\midrule
CenterPoint-Voxel~\cite{yin2021center}  &         -      & L        & 56.4   & 64.8  \\
CenterPoint-Pillar &       -        & L        & 50.3 & 60.2  \\ 
\midrule
FCOS3D~\cite{wang2021fcos3d}            & 900$\times$1600      & C        & 29.5 & 37.2  \\
DETR3D~\cite{wang2022detr3d}             & 900$\times$1600      & C        & 30.3 & 37.4  \\
BEVDet-R50~\cite{huang2021bevdet}         & 256$\times$704       & C        & 28.6 & 37.2  \\
BEVDet-Base        & 512$\times$1408      & C        & 34.9 & 41.7  \\
PETR-R50~\cite{liu2022petr}           & 384$\times$1056      & C        & 31.3 & 38.1  \\
PETR-R101          & 512$\times$1408      & C        & 35.7 & 42.1  \\
PETR-Tiny          & 512$\times$1408      & C        & 36.1 & 43.1  \\
BEVDet4D-Tiny~\cite{huang2022bevdet4d}      & 256$\times$704       & C        & 32.3 & 45.3  \\
BEVDet4D-Base      & 640$\times$1600      & C        & 39.6 & 51.5  \\
BEVFormer-S~\cite{li2022bevformer}        &       -        & C        & 37.5 & 44.8  \\
BEVDepth-R50~\cite{li2022bevdepth}        & 256$\times$704       & C        & 35.9 & 48.0  \\
BEVDepth-ConvNext       & 512$\times$1408      & C        & 46.2 & 55.8  \\
\midrule
BEVStereo-R50        & 256$\times$704       & C        & 37.6 & 49.7  \\
BEVStereo-ConvNext   & 512$\times$1408       & C        & 47.8 & 57.5  \\
\bottomrule
\end{tabular}}
\caption{Comparison on the nuScenes \emph{val} set. L denotes LiDAR and C denotes camera.}
\label{tab:val}
\end{table}

\begin{figure*}[h!]
\includegraphics[width=0.9\textwidth]{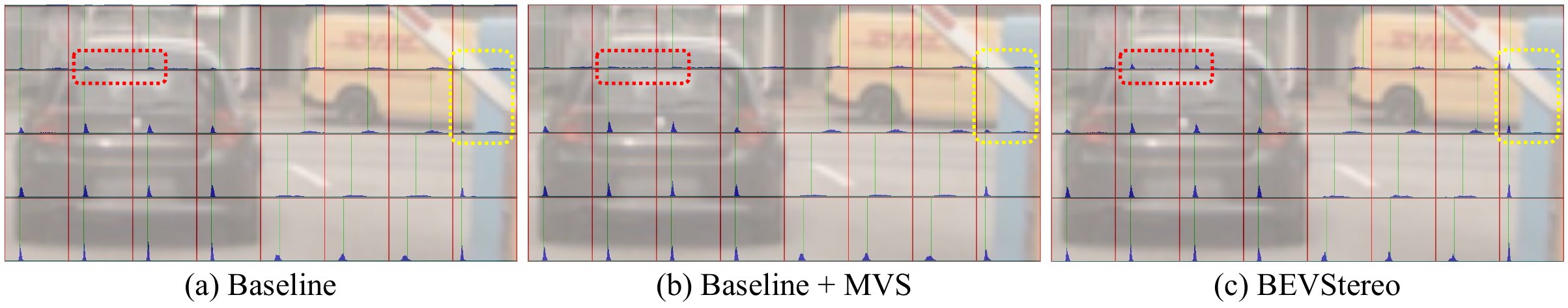}
\centering
\caption{Visualization of depth prediction. The blue area is the distribution of depth prediction, while the green line represents the depth GT produced by the point cloud. The red dotted boxes denotes the promotion of depth prediction on moving objects and the yellow dotted boxes denotes the the promotion of depth prediction on static objects.}
\label{fig:visualize_depth}
\end{figure*}


\begin{figure*}[t!]
\includegraphics[width=0.84\textwidth]{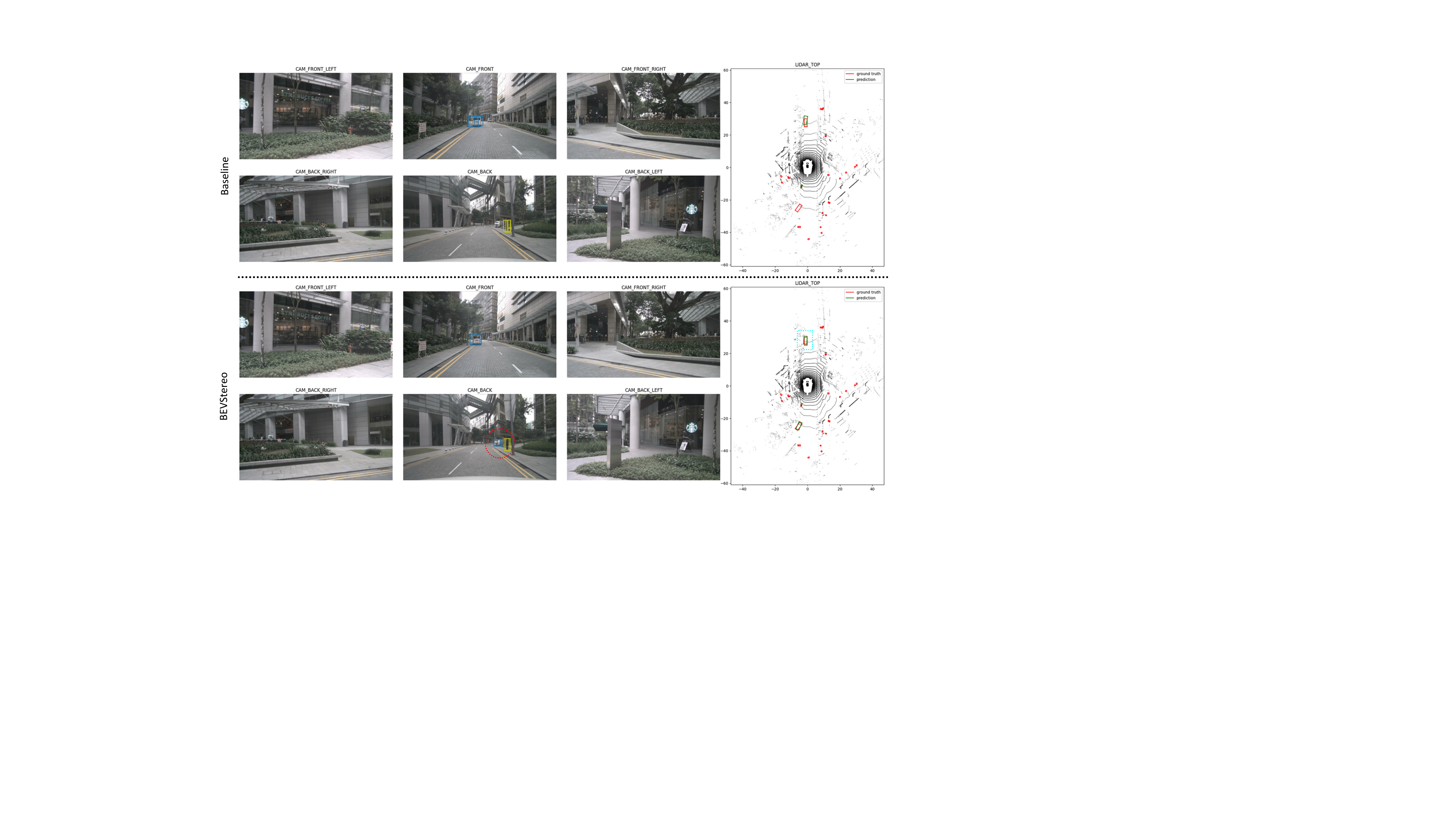}
\centering
\caption{Visualization of detection results. The blue dotted rectangle designates the object recognized by our approach is more accurate on localization, while the red dotted circle designates the object detected by BEVStereo but missed by the baseline.}
\label{fig:visualize_detection}
\end{figure*}

\subsection{Ablation Study}
\paragraph{Iteration of $\mu$ and $\sigma$}
We conduct various experiments during the inference stage by modifying the number of iterations just to verify the function of iterating $\mu$ and $\sigma$. As illustrated in Tab.~\ref{tab:iteration}, the detection results improve as the number of iterations grows.
\paragraph{Weight Net}
We run the experiment under identical conditions without Weight Net to assess its validity. Weight Net promotes the detection results, as shown in Tab.~\ref{tab:performance-analysis-1}. 

\paragraph{Size-aware Circle NMS}
We compare BEVStereo with the size-aware circle NMS to BEVStereo with the conventional circle NMS as our baseline. They are subjected to class-aware and class-agnostic procedures in order to test the validity of size-aware circle NMS.

As shown in Tab.~\ref{tab:circlenms}, our size-aware circle NMS improves on the matrices of mAP, mATE, and NDS when using class-aware NMS. The traditional distance-based circle NMS has completely lost its capacity to suppress under class-agnostic circumstance, while our size-aware circle NMS continues to function well.

\paragraph{Efficient Voxel Pooling v2}
In the previous version of Efficient Voxel Pooling~\cite{li2022bevdepth}, threads within the same warp access memory discontinuously, leading to more memory transactions, which results in poor performance. We enhance Efficient Voxel Pooling by improving the way threads are mapped, as illustrated in Fig.~\ref{fig:mapping}. For each block, we employ 32 and 4 threads on the x and y axes. First, 128 point coordinates are loaded into shared memory by all the threads in one block. Then, one point feature at a time is processed by each warp. According to the point coordinates, the point feature is atomically accumulated to the matching BEV feature. The 128 point features are processed round robin by four warps in a block till they are finished. In this manner, performance-limiting memory transactions from the L2 cache and global memory are diminished.

We compare the latency of Efficient Voxel Pooling v1 and Efficient Voxel Pooling v2 using various resolutions. Efficient Voxel Pooling v2 is able to reduce the latency up to 40\%.

\subsection{Visualization}
As illustrated in Fig.~\ref{fig:visualize_depth}, we can find that BEVStereo has the ability to promote the accuracy of depth estimation on both moving and static objects. We also visualize the detection results, as shown in Fig.~\ref{fig:visualize_detection} which also demonstrates the performance promotion brought by BEVStereo.

\subsection{Benchmark Result}
We compare BEVStereo with other state-of-the-art methods like CenterPoint ~\cite{yin2021center}, FCOS3D~\cite{wang2021fcos3d}, DETR3D~\cite{wang2022detr3d}, BEVDet~\cite{huang2021bevdet}, PETR~\cite{liu2022petr}, BEVDet4D~\cite{huang2022bevdet4d} and BEVFormer~\cite{li2022bevformer}. We evaluate our BEVStereo on the nuScenes \emph{test} and \emph{val} set. As shown in Tab.~\ref{tab:test} and Tab.~\ref{tab:val}, BEVStereo achieves the highest score of camera-based methods on both mAP and NDS.


\section{Conclusion}
In this paper, a novel multi-view 3D object detector is proposed, namely BEVStereo. BEVStereo improves performance without significantly increasing memory usage by applying dynamic temporal stereo technique to create temporal stereo. Some complex scenarios that other stereo-based approaches cannot handle can be resolved by our method. In addition, we propose size-aware circle NMS, which takes the size of boxes into account while avoiding the laborious computation of rotated IoU. Under both class-aware and class-agnostic circumstances, our size-aware circle NMS performs satisfactorily. Last but not least, we present Efficient Voxel Pooling v2, which speeds up voxel pooling by improving the efficiency of memory accesses.

\section{Acknowledgements}

Throughout the process of developing BEVStereo, I have received a great deal of guidance and assistance. I would like to thank Haotian Zhang, Yuefeng Wu and Tai Wang for their wonderful collaboration and patient support.
\bibliography{aaai23}
\end{document}